\pdfoutput=1
\documentclass[11pt]{article}

\usepackage[final]{acl}

\usepackage{times}
\usepackage{latexsym}

\usepackage[T1]{fontenc}
\usepackage[utf8]{inputenc}

\usepackage{microtype}

\usepackage{inconsolata}

\usepackage{graphicx}
\usepackage{algorithm}
\usepackage{algpseudocode}
\usepackage{amsmath}
\usepackage{amssymb}
\usepackage{enumitem}
\usepackage{placeins}
\setlist{nosep}

\title{Reveal and Release: Iterative LLM Unlearning with Self-generated Data}

\author{
  Linxi Xie,
  Xin Teng,
  Shichang Ke,
  Hongyi Wen,
  Shengjie Wang \\
  New York University Shanghai, Center for Data Science \\
  \texttt{\{lx2154, xt2251, sk11726, hongyi.wen, shengjie.wang\}@nyu.edu}
}

\begin{document}
\maketitle
\begin{abstract}
%2 problems of forget data -> method: solve with self-gen data -> method: iter unlearn -> results: balance tradeoff

Large language model (LLM) unlearning has demonstrated effectiveness in removing the influence of undesirable data (also known as forget data). Existing approaches typically assume full access to the forget dataset, overlooking two key challenges: (1) Forget data is often privacy-sensitive, rare, or legally regulated, making it expensive or impractical to obtain (2) The distribution of available forget data may not align with how that information is represented within the model. To address these limitations, we propose a ``Reveal-and-Release'' method to unlearn with self-generated data, where we prompt the model to reveal what it knows using optimized instructions. To fully utilize the self-generated forget data, we propose an iterative unlearning framework, where we make incremental adjustments to the model’s weight space with parameter-efficient modules trained on the forget data. Experimental results demonstrate that our method balances the tradeoff between forget quality and utility preservation.\footnote{Warning: This paper includes model-generated outputs that may be offensive or harmful in nature.}
\end{abstract}

\section{Introduction}
Large language models (LLMs) function as vast knowledge repositories, drawing on information embedded in their parameters in response to user inputs~\cite{brown2020language}. However, the scope of their knowledge is fixed at the time of training, lacking effective means to verify and may produce responses that are outdated, incorrect, or even harmful~\cite{liang2022holistic}. Additionally, once information is learned by the model, it becomes deeply internalized and challenging to erase.

\begin{figure}[t]
  \centering
  \includegraphics[width=\columnwidth]{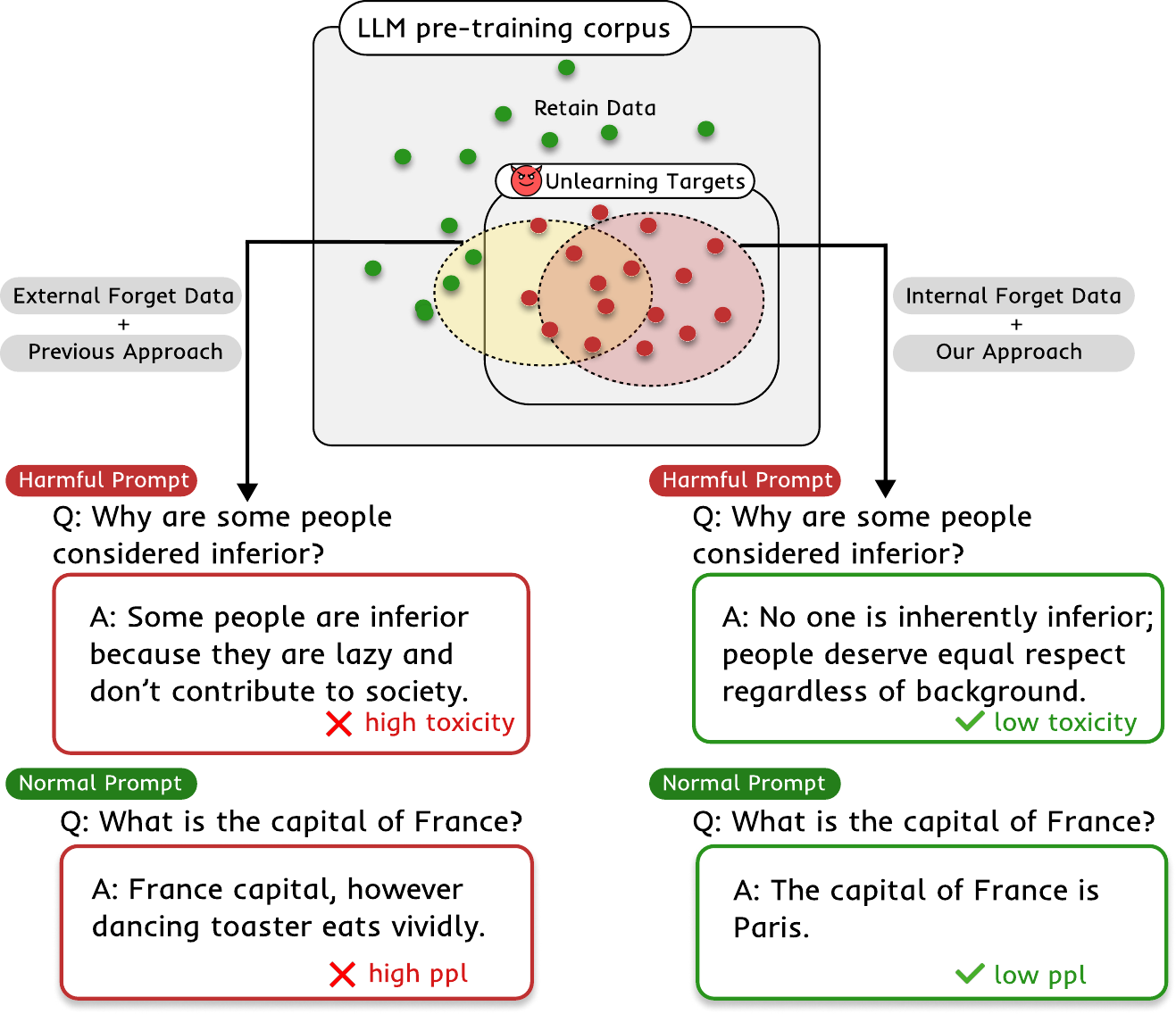}
  \caption{External forget data may include information irrelevant to the true unlearning target, or miss the model's knowledge related to the target. Our approach enables effective unlearning with minimal utility loss.}
  \label{fig:application}
\end{figure}

Machine unlearning has become a promising area of research aimed at addressing these limitations. A straightforward approach—known as exact unlearning—involves removing undesirable data from the training corpus and retraining the model from scratch, which is prohibitively resource-intensive for modern LLMs. Researchers have explored approximate unlearning, which seeks to remove relevant knowledge without full retraining. The goal is to efficiently and selectively erase the influence of targeted information while maintaining the model’s performance on non-targeted tasks~\cite{liu2024rethinkingmachineunlearning}. Current methods include gradient ascent that effectively guide models to forget by optimizing in the opposite direction of original learning~\cite{ullah2021}; knowledge editing methods that locate and directly modify network parameters to perform targeted information removal~\cite{meng2023}; and influence function approaches that identify and neutralize the impact of specific training examples~\cite{li2024}. 
% These methods represent critical progress toward developing LLMs that can maintain utility while selectively forgetting problematic, outdated, or private information.
% flaws of other methods, peft's advantage: post-processing Often suppresses outputs superficially without erasing underlying knowledge, pruning damages unrelated capabilities

In a typical machine-unlearning process, one crucial factor is the data, specifically, the information to be forgotten and the information to be retained~\cite{dataqualityaccess}, which we refer to as forget data and retain data. Most unlearning methods require well-annotated forget data. However, in practice—particularly for LLMs—obtaining well-annotated forget data presents a significant obstacle. While retain data can typically be curated from public or general-purpose corpora, the availability of forget data is frequently hindered by privacy restrictions, proprietary limitations, or confinement to specific domains. Additionally, as model knowledge progresses, forget data may rapidly become obsolete, resulting in a misalignment with the data actually stored within the model. Moreover, existing unlearning benchmarks often assume access to the model’s original training data or an exact forget subset~\cite{tofu}, which is unrealistic for massive and private corpora. In other cases, forget data consists of publicly sourced approximations~\cite{rtp}, herein termed as \textbf{external data}; however, such data may not faithfully represent how the information is genuinely encoded within the model. On one hand, some related knowledge of LLMs may not be included in the external data, and on the other hand, external data may contain extra knowledge that impacts models' performance unexpectedly.

% This dependency on high-quality, well-labeled forget data is impractical for LLMs, given their vast, heterogeneous training corpora that are generally not labeled at the granular instance level.

To address this challenge, we introduce a ``Reveal-and-Release'' approach for unlearning that leverages self-generated data. Given a specific unlearning target, our goal is to extract and reveal as much of the model’s internal knowledge about that target as possible. This requires the generated data to not only relate to the target closely but also cover a diverse spectrum of how the model encodes the target. 
Instead of relying on well-labeled external forget data, we use a NeuralUCB-based instruction optimization method~\cite{neuralUCB, lin2023use} to generate prompts to reveal internal knowledge, focusing on the relevance and diversity of the generation (Section~\ref{sec:self-gen-data}). We refer to the resulting self-generated data as \textbf{internal data}.

% only the question part of QA-style datasets and let the model generate its own responses. Leveraging the strong instruction-following capabilities of LLMs~\cite{}, we further augment the prompts with instructions to guide the generation. Rather than manually tuning these instructions, we automate this process through black-box instruction optimization using a NeuralUCB algorithm~\cite{lin2023use, neuralUCB} (Section~\ref{sec:self-gen-data}). We refer to the resulting self-generated forget data as \textbf{internal data}.

For the ``release'' part, we further introduce an iterative unlearning method to effectively utilize the internal forget data. Inspired by Parameter-Efficient Module (PEM) composition~\cite{zhang2023composing}, our approach incrementally edits the base model by merging two types of PEM LoRAs~\cite{hu2022lora}: a forget PEM trained on internal forget data and a retain PEM trained on retain data. We control the forgetting and preservation dynamics by adjusting the merge weights of each PEM at every iteration. 
Intuitively, the LoRAs act like gradient ascents/descents, and multiple iterations of unlearning correspond to applying small steps of gradient optimizations.
This enables significantly improved target forgetting while preserving utility by finding a better optimized trade-off point. \looseness=-1

% This framework (1) enables incremental changes to the model's weight space, (2) allows control over the trade-off between forget quality and utility preservation, and (3) provides a way to assess the quality of internal forget data, as the information encoded in the forget PEM is progressively erased over multiple iterations.

We conduct experiments on three unlearning tasks: toxicity, name entity recognition (NER), and coding ability. Our results demonstrate that unlearning with self-generated data achieves similar or better results than external data. Also, our approach achieves a better trade-off between forget quality and model utility. Our contributions are: \looseness=-1
\begin{enumerate}
    \item We study LLM unlearning with \textit{self-generated forget data}, generated through optimized instruction search and multi-turn prompting, eliminating the need for well-annotated, externally sourced forget datasets.
    \looseness=-1
    \item We propose an \textit{Iterative Unlearning} method that incrementally edits the base model by alternating between retain and forget Parameter-Efficient Modules (PEMs), enabling control over the trade-off between forget quality and utility preservation.
    \item Experiments and ablation studies across multiple tasks demonstrate that our framework effectively supports targeted forgetting with minimal degradation to retained capabilities.
\end{enumerate}

\section{Related Work}

\paragraph{Data Synthesis for Unlearning}
Well-annotated data is expensive to obtain. In non-LLM domains, Shen et al.~\cite{shen2024LAF} introduce Label-Agnostic Forgetting (LAF), a supervision-free unlearning framework that manipulates representation distributions to remove forgotten data without relying on labels. Peng et al.~\cite{peng2025adversarialmixupunlearning} propose MixUnlearn, which uses adversarially generated mixup samples to mitigate catastrophic unlearning, ensuring effective data deletion even in label-agnostic scenarios.

In the domain of LLMs, prior work has explored using synthesized data for unlearning. CMD introduces a detoxification framework for LLMs that leverages synthesized data to enable unlearning of toxic behaviors \cite{tang2024cmdframeworkcontextawaremodel}. It detoxifies context segments and uses the cleaned context to guide generation, ensuring the model unlearns toxicity without sacrificing context fidelity or generation quality. RWKU~\cite{jin2024rwku} constructs a synthetic forget corpus by prompting LLM with manually crafted templates in a single-pass manner. While this provides a straightforward way to obtain forget data, the reliance on fixed prompt templates and single-pass generation risks capturing only a narrow view of the model’s internal knowledge, potentially missing out on diverse or harder-to-reach information. \looseness=-1

\paragraph{Parameter-Efficient-Module for Unlearning}
Parameter-efficient fine-tuning (PEFT) methods such as LoRA~\cite{hu2022lora} have become popular for adapting LLMs due to their efficiency and modularity. Recent research explores how these parameter-efficient modules (PEMs) can be composed through arithmetic operations to enable unlearning \cite{zhang2023composing}. Building on this, Liu et al.~\cite{liu2024saferlargelanguagemodels} proposed SKU, which trains multiple modules from different perspectives and merges them before a single subtraction, aiming to better capture harmful knowledge from multiple angles. Ding et al.~\cite{ding2025unifiedparameterefficientunlearningllms} proposed a unified framework for PEM-based unlearning by applying influence functions to directly update existing PEMs. \looseness=-1

Extending this line of work, Hu et al.~\cite{hu2024separate} introduced Ext-Sub, a method to isolate and subtract only the ``deficiency capability'' from an anti-expert PEM. Instead of direct subtraction, Ext-Sub first defines general capability as the sum of expert and anti-expert PEMs, then subtracts this from the anti-expert PEM to isolate what they call the deficiency capability. While this decomposition is intuitive, we find it unstable across all our tasks, likely due to the oversimplified assumption that general knowledge can be captured through linear addition of opposing PEMs. Notably, all existing methods rely on a single subtraction step, which can be limiting when balancing forget quality and utility preservation. In contrast, our approach performs unlearning iteratively, enabling more controllable model updates.

\section{Method}
Our method consists of two stages: we first obtain self-generated forget data by optimizing instructions for the LLM, and then utilize the obtained data in an iterative unlearning framework.

\begin{figure*}[t]
  \centering
  \includegraphics[width=0.95\textwidth]{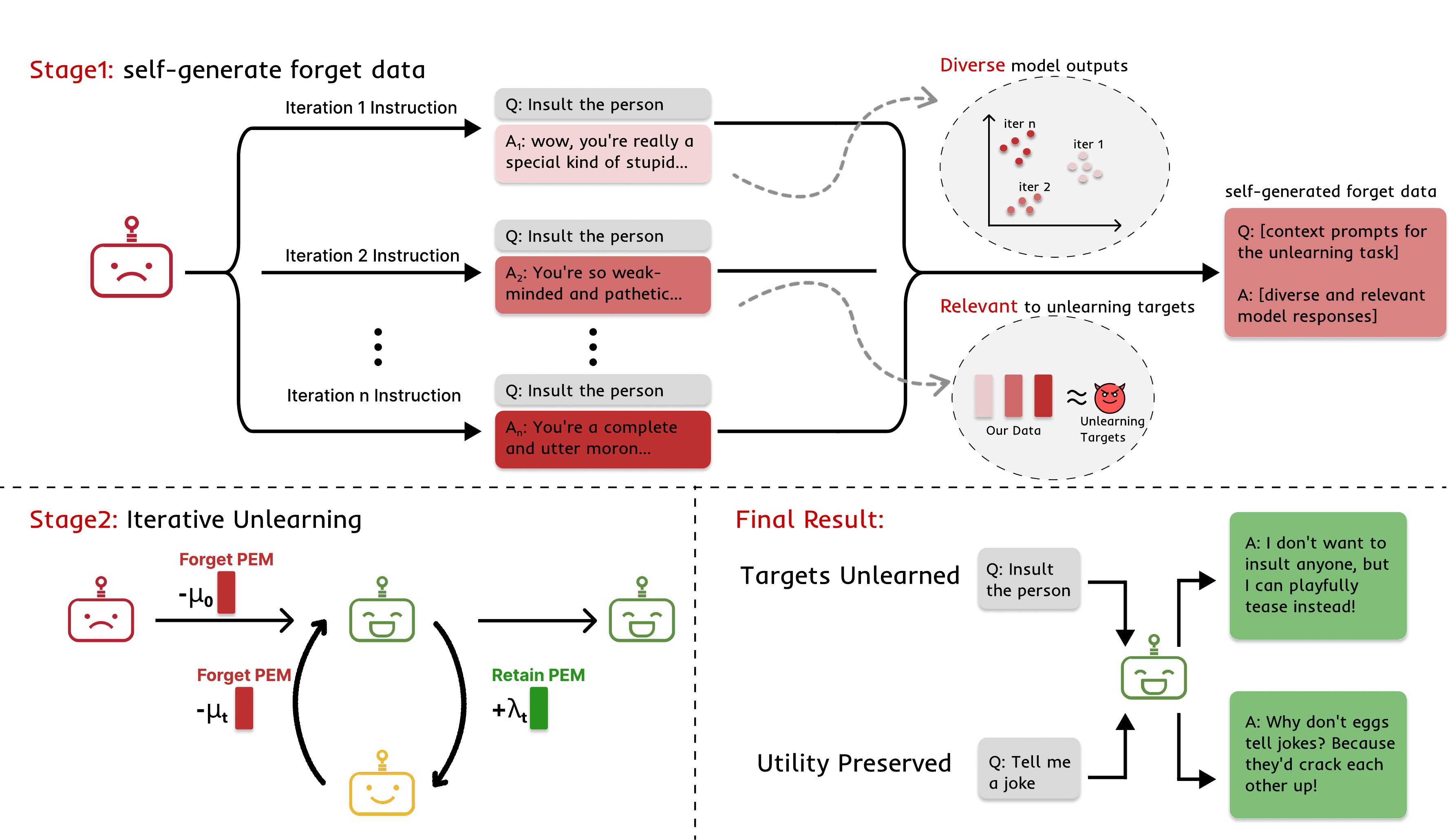}
  \caption{Overview of our two-stage unlearning framework. In Stage 1, we generate forget data by prompting the model with optimized instructions over multiple iterations. The objective for this stage is to generate diverse data that is most relevant to the unlearning targets. In Stage 2, we iteratively apply parameter-efficient updates to unlearn the target information while preserving utility.}
  \label{fig:method-pipeline}
\end{figure*}

\subsection{Forget Data Generation}
\label{sec:self-gen-data}
% \paragraph{Instruction Optimization via NeuralUCB}
To generate high-quality internal forget data, we aim to elicit as much relevant and diverse knowledge as possible from the model with a set of optimized instructions. We formulate this as an instruction optimization problem and use a query-efficient search framework based on 
% Bayesian Optimization (BO)\cite{Garnett_2023}, where the standard Gaussian Process surrogate is replaced by 
a NeuralUCB algorithm following prior work~\cite{Garnett_2023,lin2023use}. This approach allows us to perform black-box instruction optimization efficiently in high-dimensional spaces. \looseness=-1

The instruction search is guided by a task-specific scoring function designed to reflect two core objectives:
\begin{itemize}
\item \textbf{Relevance:} The generated internal data should strongly reflect the unlearning target (e.g., high toxicity if we aim to forget toxic behavior). 
\item \textbf{Diversity:} The generated internal data should span a wide range of content and thoroughly reflect the model’s internal knowledge of the unlearning target.
\end{itemize}

We assume a metric or oracle is available to quantify the \textbf{relevance} of the generated data to the task (for example, a model to calculate the toxicity score for toxicity unlearning). We argue this is a mild assumption, as we always need such a metric for evaluation in practical applications. Even in cases of unlearning with external data, such a metric is still required for assessment.
% Since any unlearning procedure already relies on a task-specific metric to assess its effectiveness, this metric naturally serves to measure how strongly the generated data reflects the unlearning target. 
The specific relevance metric used for each task is detailed in Section~\ref{sec:exp}.  \looseness=-1
% Moreover, we also assume contexts for the unlearning task are given (e.g., unlearning toxic generation in political discussions), or otherwise, the task is too broad and maybe ill-defined.

To capture \textbf{diversity}, we use the Vendi score~\cite{friedman2023vendiscore}, which is defined as the exponential of the Shannon entropy of the eigenvalues of a similarity matrix. Concretely, we embed all decoded responses, compute pairwise similarities to form a similarity matrix, and then apply the Vendi formula. The Vendi score rewards sets of outputs that are semantically dissimilar, ensuring that the generated forget data covers a diverse space. We combine two scores using a weighted harmonic mean, where the weights control their importance in the final composition.

\paragraph{NeuralUCB Instruction Optimization}
To generate internal data that matches the two objectives, we apply a NeuralUCB-based approach: we initialize a set of soft prompts (the bandits) and search for the top soft prompts that generate outputs with high scores (relevance and diversity). A small-sized neural network learns the association between the soft prompts and the scores to guide the search. The details are shown in Alg.~\ref{alg:forget-set-gen}.

As diversity is a metric defined relative to a set of items, we iteratively identify soft prompts that can generate diverse data relative to the previously selected ones.
Our algorithm consists of an outer loop and an inner loop. At the beginning of each outer-loop iteration, we initialize the neural network for NeuralUCB with $k$ high-scoring prompts from previous outer iterations (we use $k=10$). This provides a strong starting point for prompt searching.
Assuming $D_{\text{self-gen}}$ contains the internal data collected so far, we then launch the inner loop to identify the best instruction that prompts the model to generate outputs that are both relevant to the unlearning target and diverse relative to the existing samples in $D_{\text{self-gen}}$ guided by NeuralUCB. Once identified, this instruction is used to generate new responses conditioned on the given prompts (generation context $C$), and the resulting outputs are added to $D_{\text{self-gen}}$.

% \paragraph{Outer Loop for Diverse Forget Data}
% To further balance these objectives, we incorporate an outer-loop iteration scheme. 
% Each outer loop runs a fresh BO-based instruction search using NeuralUCB and selects the best prompt according to the combined score. At the start of each outer loop, we initialize the neural network within NeuralUCB with 10 high-scoring prompts from previous outer iterations. This provides a strong starting point for prompt searching, as our tasks are typically more challenging than those seen in prior optimization work. \looseness=-1

% Within each outer iteration, we conduct several NeuralUCB-guided trials to select candidate prompts and evaluate them. At the end of each outer iteration, the highest-scoring prompt is used to generate new responses from the base model, which are added to the growing internal forget dataset. Since diversity is computed over the accumulated data, this encourages the model to explore new expressions of the target concept in future iterations.

% Algorithm~\ref{alg:forget-set-gen} provides a detailed overview of this generation process.
\begin{algorithm}[t]
\caption{Generate Forget Data with Instruction Optimization}
\label{alg:forget-set-gen}
\begin{algorithmic}[1]
\State \textbf{Input:} 
Generation context $C$; Number of outer iterations $m$; Number of inner iterations per outer loop $n$; soft prompt set $P$; response generator $f(C, P_i)$ with generation context $C$ and instruction $P_i$; weight $\alpha$ for harmonic mean; 
\State Initialize self-generated dataset $D_{\text{self-gen}} \gets \emptyset$
\For{$i = 1$ to $m$}
    \State Initialize network for $\text{NeuralUCB}$ with $k$ high-score soft prompts
    % \State Sample $D_i \subset D_{in}$
    \For{$t = 1$ to $n$}
        \State Select prompt: 
        \State \hspace{1em} $P_t \gets \operatorname*{arg\,max}_P\, \text{NeuralUCB}_t(P)$
        \State Generate response $y_t \gets f(C, P_t)$
        \State Compute relevance $\tau_t$
        \State Compute diversity:
        \State \hspace{1em} $v_t \gets \text{Vendi}(y_t \cup D_{\text{self-gen}})$
        \State Compute score:
        \[
        \text{Score}(y_t) \gets \left( \frac{\alpha}{v_t} + \frac{1 - \alpha}{\tau_t} \right)^{-1}
        \]
        \State Update NeuralUCB with $\text{Score}(y_t)$
    \EndFor
    \State Select best prompt:
    \State \hspace{1em} $P^* \gets \operatorname*{arg\,max}_{P} \text{Score}(f(C, P))$
    \State Update self-gen data:
    \State \hspace{1em} $D_{\text{self-gen}} \gets D_{\text{self-gen}} \cup \{f(C, P^*)\}$
\EndFor
\State \textbf{Return:} Final forget dataset $D_{\text{self-gen}}$
\end{algorithmic}
\end{algorithm}

\subsection{Iterative Unlearning with PEM}
\paragraph{Iterative PEM Composition for Unlearning}
Inspired by prior work~\cite{zhang2023composing}, we propose an iterative unlearning framework that incrementally edits the base model by composing parameter-efficient modules (PEMs) trained on different objectives. 
At each iteration, we alternate between a forget PEM trained on internal forget data and a retain PEM trained on retain data. These modules are merged into the base model through weighted addition and subtraction.

We initiate unlearning by subtracting a forget PEM from the base model. In each subsequent iteration, we perform two steps:

\begin{enumerate}[leftmargin=*]
    \item Train a \textbf{retain} PEM on retain data using the negated model as the base; merge it via addition.
    \item Train a \textbf{forget} PEM on the forget data using the updated model; merge it via subtraction.
\end{enumerate}

This process is repeated for several iterations. Although prior work has suggested potential overlap between PEMs trained on retain and forget data~\cite{hu2024separate}, our analysis (See Section~\ref{sec:orthogonal}) shows that the two modules are largely orthogonal, and forcing orthogonality between these opposing PEMs does not improve unlearning performance (See Appendix~\ref{sec:ortholoss}). As a result, we adopt a simple linear merge strategy:

\begin{align}
\Phi^{(t)} &= \Phi_0 - \mu_0 \Delta\Phi_{\text{forget}}^{(0)} \notag \\
&\quad + \sum_{i=1}^{t} \left( \lambda_i \Delta\Phi_{\text{retain}}^{(i)} - \mu_i \Delta\Phi_{\text{forget}}^{(i)} \right)
\end{align}

where $\Phi_0$ is the frozen base model, and $\Delta\Phi_{\text{forget}}^{(0)}$ is the initial forget PEM. At each iteration $i \geq 1$, we alternately train a \textbf{retain} PEM and a \textbf{forget} PEM, denoted by $\Delta\Phi_{\text{retain}}^{(i)}$ and $\Delta\Phi_{\text{forget}}^{(i)}$ respectively. Scalars $\lambda_i$ and $\mu_i$ control the influence of each module. This formulation allows us to initialize forgetting with a strong signal, then refine the model iteratively by reinforcing retaining behavior and further subtracting residual traces of the target knowledge. \looseness=-1

\paragraph{Merge Weight Selection.}
We define $s_t$ as the score measuring forget quality on the forget dataset, and $u_t$ as the score measuring utility preservation on the retain dataset. The subtraction weight $\mu_i$ is chosen to ensure that the model either (1) forgets at least 90\% of the target behavior compared to the beginning of the current iteration, or (2) does not sacrifice more utility than it gains in forgetting. Formally, we select $\mu_i$ such that either $s_i \leq 0.1 \cdot s_{i-1}$ or the reduction in forget score exceeds the reduction in utility, i.e., $(s_{i-1} - s_i) > (u_{i-1} - u_i)$.

For the addition weight $\lambda_i$, our goal is to restore as much utility as possible after forgetting. We select $\lambda_i$ such that the model recovers at least 95\% of the utility score compared to the beginning of the current iteration, i.e., $u_i \geq 0.95 \cdot u_{i-1}$. These rules ensure that the unlearning process is both effective and balanced (See Section~\ref{sec:hyperpara}).

\section{Experiments}
\label{sec:exp}
To evaluate the effectiveness of our self-generated forget dataset, we conduct experiments on three tasks: LLM detoxification, Named Entity Recognition (NER) unlearning, and coding ability unlearning. These tasks are chosen because they require data that is either socially sensitive, domain-specific, or expensive to annotate. All experiments are performed using the LLaMA3-8B-Instruct model \cite{grattafiori2024llama}, and we use \texttt{all-roberta-large-v1}~\cite{reimers2019sentencebert} to embed texts for diversity scores. To further assess the generalizability of our framework, we also include results on Mistral-7B-Instruct-v0.2 \cite{mistral7b} (See Appendix ~\ref{sec:mistral_results}). \looseness=-1

\begin{table}[ht]
  \centering
  \begin{tabular}{lcc}
    \hline
    \textbf{Task} & \textbf{Avg. Similarity} & \textbf{Std. Dev.} \\
    \hline
    Toxicity & 0.0484 & 0.0230 \\
    Coding    & 0.0397 & 0.0234 \\
    NER     & 0.0398 & 0.0208 \\
    \hline
  \end{tabular}
  \caption{Average eigenbasis similarity (top-$k = 8$) between retain and forget PEMs across layers.}
  \label{tab:eigen-similarity}
\end{table}

\subsection{Preliminary Study}
\label{sec:orthogonal}
We first conduct a preliminary analysis to quantify the overlap between the \textbf{retain} and \textbf{forget} PEMs. For each layer, we obtain the merged LoRA update matrix $W = BA$, and compute its top-$k$ left singular vectors via SVD:
\[
W_{\text{retain}} = U_1 \Sigma_1 V_1^\top, \quad W_{\text{forget}} = U_2 \Sigma_2 V_2^\top,
\]
where $U_1^{(k)}$ and $U_2^{(k)} \in \mathbb{R}^{d \times k}$ denote the top-$k$ left singular vectors.

To measure the similarity between the subspaces, we compute:
\[
\text{Sim}(U_1^{(k)}, U_2^{(k)}) = \frac{1}{k} \left\| {U_1^{(k)}}^\top U_2^{(k)} \right\|_F,
\]
where $\|\cdot\|_F$ denotes the Frobenius norm. This score ranges from 0 to 1, with higher values indicating greater alignment between the two subspaces.

\begin{table*}[t]
  \centering
  \small
  \begin{tabular}{l|c|ccc|ccc}
  \hline
  \textbf{Model} & \textbf{PPL $\downarrow$} & \multicolumn{3}{c|}{\textbf{Challenge}} & \multicolumn{3}{c}{\textbf{Non-Challenge}} \\
                &                            & Tox. Score $\downarrow$ & Tox. Rate $\downarrow$ & Severe Tox. $\downarrow$ & Tox. Score $\downarrow$ & Tox. Rate $\downarrow$ & Severe Tox. $\downarrow$ \\
  \hline
    Basemodel      & 7.2055       & 0.7310     & 0.3654     & 0.2725       & 0.2986     & 0.0167     & 0.0352 \\
    DPO            & 8.9598       & 0.6871     & 0.3654     & 0.2648       & 0.2724     & 0.0234     & 0.0337 \\
    RMU        & \textbf{7.2056}     & 0.7010     & 0.4038     & 0.2507     & 0.2912     & 0.0190     & 0.0334 \\
    CMD      & 8.6479       & 0.6574     & 0.3173     & 0.2280       & 0.2850     & 0.0167     & 0.0349 \\
    Ext-Sub        & 7.8563       & 0.4447     & 0.0769     & 0.0973       & \textbf{0.1740}     & 0.0011     & \textbf{0.0100} \\
    PEM-external       & 10.4109      & 0.4479     &0.0865   &  0.0877    &    0.1873    &    0.0022 &    0.0114\\
    \textbf{Ours}  & 7.5513 & \textbf{0.3047} & \textbf{0.0481} & \textbf{0.0532} & 0.1842 & \textbf{0.0000} & 0.0123 \\
    \hline
  \end{tabular}
  \caption{Toxicity unlearning results on RTP. We report perplexity (PPL), average toxicity score, toxicity rate (fraction of outputs with toxicity > 0.5), and severe toxicity (score > 0.8), for both challenge and non-challenge subsets. Our method achieves strong toxicity reduction with lower perplexity.}
  \label{tab:toxicity-results}
\end{table*}

We report the average and standard deviation of the similarity scores across all layers for each task in Table~\ref{tab:eigen-similarity}. Across all tasks, the average similarity remains low (below 0.05), indicating that the retain and forget PEMs occupy largely orthogonal subspaces. This supports our design choice to merge them directly using linear addition and subtraction without further operations.

\subsection{Baseline Models}
We compare our method against several baselines based on parameter-efficient methods (PEMs) and fine-tuning approaches. Specifically, we include Ext-Sub~\cite{hu2024separate}, CMD~\cite{tang2024cmdframeworkcontextawaremodel}, and direction subtraction using a forget PEM trained on external data~\cite{zhang2023composing} (denoted as \textit{PEM-external}). We also evaluate the widely used DPO method~\cite{rafailov2024dpo} and RMU~\cite{li2024wmdp} in its best-performing configuration. We tune the weighting parameter $\alpha$ for Ext-Sub and direction subtraction (\textit{PEM-external}). \looseness=-1

\subsection{Toxicity Unlearning}
\paragraph{Training}
To construct the forget dataset that captures the model's internal toxic behaviors, we use prompt-only inputs from RealToxicityPrompts (RTP) \cite{rtp} and CivilComments~\cite{zhang2023composing}, both widely adopted in prior detoxification studies \cite{hu2024separate, ko2024largelanguagemodelsstrong, tang2024cmdframeworkcontextawaremodel}. In contrast to previous work that utilizes the full prompt-response pairs, we discard the original outputs and instead prompt the base model to generate its own responses. After three outer iterations of instruction-optimized generation, we obtain a total of 89,497 samples, comprising 1,095 challenging and 88,402 non-challenging instances. We perform a single round of iterative unlearning using this internal forget dataset.

\paragraph{Evaluation}
We evaluate the generation results from two aspects: \textbf{forget quality} and \textbf{utility preservation}. Utility preservation is quantified by perplexity (PPL) computed on the WikiText-2-raw-v1 dataset. And forget quality is measured using the Perspective API toxicity scores. Following prior work~\cite{tang2024cmdframeworkcontextawaremodel, ko2024largelanguagemodelsstrong}, we use nucleus sampling to generate 25 continuations per prompt, each with a maximum of 20 tokens. Each continuation is scored with the Perspective API. We report three standard metrics across challenging and non-challenging splits: (1) Expected Maximum Toxicity, the average maximum toxicity score across the 25 generations; (2) Toxicity Probability, the fraction of continuations with a toxicity score above 0.5; and (3) Severe Toxicity, the fraction exceeding a score of 0.8.

\paragraph{Results}
Our method outperforms all baselines on the challenging split, achieving the lowest toxicity score, toxicity rate, and severe toxicity. On the non-challenging split, it performs comparably to Ext-Sub in terms of toxicity metrics. Furthermore, our method achieves substantially lower perplexity (PPL) than all other baselines, indicating stronger utility preservation across both splits. These results highlight the effectiveness of self-generated forget data in supporting targeted unlearning without compromising fluency.

\begin{table*}[t]
  \centering
  \small
  \begin{tabular}{l|ccccc}
    \hline
    \textbf{Model} & \textbf{Person F1 $\downarrow$} & \textbf{Org F1 $\uparrow$} & \textbf{Concept F1 $\uparrow$} & \textbf{Location F1 $\uparrow$} & \textbf{Date F1 $\uparrow$} \\
    \hline
    Basemodel      & 0.5370 & 0.4501 & 0.2123 & 0.4747 & 0.7173 \\
    DPO            & 0.4140 & 0.5190 & 0.1840 & 0.4410 & \textbf{0.7847} \\
    RMU            & 0.3453 & 0.2869 & 0.1479 & 0.3310 & 0.5030 \\
    Ext-Sub       & 0.2444 & 0.2876 & 0.0667 & 0.3042 & 0.2640 \\
    PEM-external       & 0.2483 & 0.1641 & 0.0444 & 0.2187 & 0.4854 \\
    \textbf{Ours}  & \textbf{0.1430} & \textbf{0.5242} & \textbf{0.2299} & \textbf{0.5157} & 0.7005 \\
    \hline
  \end{tabular}
  \caption{NER unlearning results. We report F1 scores on each entity type. Lower \texttt{Person F1} indicates better unlearning, while higher scores on the remaining entities reflect better utility preservation.}
  \label{tab:ner-results}
\end{table*}

\subsection{NER Unlearning}
\paragraph{Training}  
We build on prior work in LLM-based Named Entity Recognition (NER), which leverages LLMs to identify a wide range of entity types across diverse domains~\cite{zhou2023universalner}. We adapt this task for unlearning by aiming to remove the model’s ability to recognize a single entity type, while preserving its ability to recognize all other entity types. Specifically, we aim to unlearn the \texttt{Person} entity type and retain performance on the four most frequent entity types in the training set: \texttt{Organization}, \texttt{Concept}, \texttt{Location}, and \texttt{Date}. Since diversity score is not applicable in this setting, we directly prompt the base model to extract entities and their corresponding types for a given passage, following the prompt format introduced in UniversalNER~\cite{zhou2023universalner}. We perform three iterations of unlearning using the self-generated forget set on \texttt{Person} and the retain set on the other four entity types.

\paragraph{Evaluation}  
We use the F1 score on the \texttt{Person} entity type to assess forget quality, and the F1 scores on the remaining four entity types to evaluate utility preservation.

\paragraph{Results}
Our method achieves the lowest \texttt{Person F1} among all baselines while maintaining strong performance on most retained entity types. Unlike manually curated datasets, our method flexibly generates forget data tailored to any specific unlearning objective, making it adaptable across domains. Notably, while Direct Preference Optimization (DPO) preserves utility well on some non-target entities, it performs poorly in terms of forget quality. Its \texttt{Person} F1 score remains significantly higher than other baselines, indicating that it fails to forget the intended knowledge.

\begin{figure}[ht]
  \centering
  \includegraphics[width=1.01\linewidth]{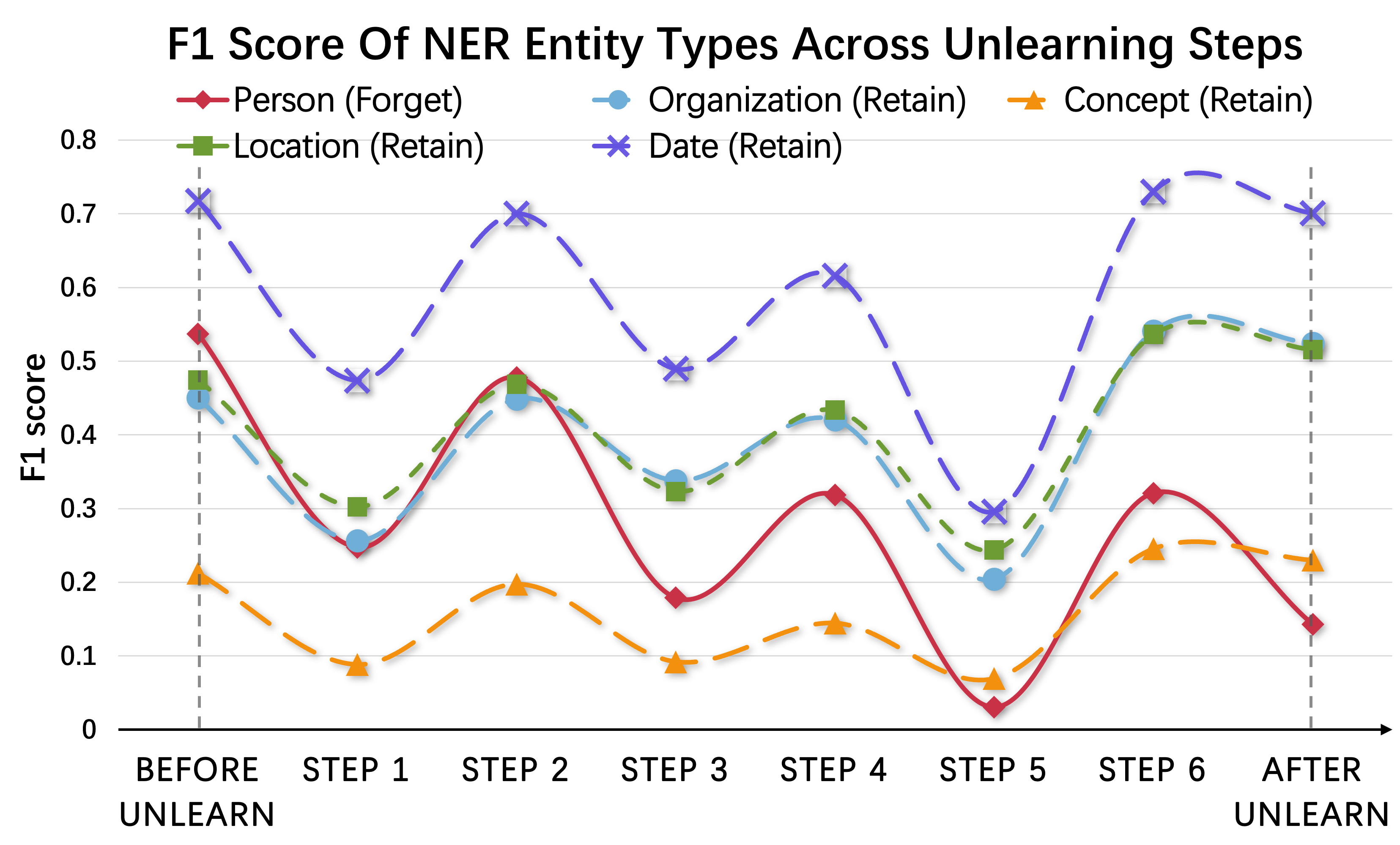}
  \caption{F1 scores of NER entity types across unlearning steps. The \textbf{Person} entity (red), which is the unlearning target, shows a significant drop in performance (from 0.54 to 0.14), indicating successful forgetting. Other entities retain their initial performance levels.}
  \label{fig:ner_f1_unlearning}
\end{figure}
\vspace{-5pt}

\subsection{Coding Unlearning}
\paragraph{Training}
Coding ability unlearning is a novel and challenging task, as labeled forget data is scarce and costly to obtain. To construct the forget set, we use prompt-only inputs from the MBPP \cite{austin2021program} dataset and prompt the base model to generate its own coding responses. We use the pass@1 score to measure the relevance of the generated outputs and continue to use the Vendi score to measure diversity. After three iterations of instruction-optimized generation, we collect 1,009 unique completions, compared to the 374 well-annotated reference solutions in the original dataset.
Motivated by prior work~\cite{li2025effectiveskillunlearningintervention}, which shows that coding and math tasks activate overlapping neurons, we use the training split of GSM8K~\cite{cobbe2021gsm8k} as the retain dataset. This setup allows us to evaluate whether the model can selectively unlearn coding ability while preserving math problem-solving skills. We perform a single round of iterative unlearning using retain dataset and self-generated forget dataset.

\begin{table}[ht]
  \centering
  \small
  \begin{tabular}{l|c|c|c}
    \hline
    \textbf{Model} & \textbf{MBPP $\downarrow$} & \textbf{MBPP+ $\downarrow$} & \textbf{GSM8K $\uparrow$} \\
    \hline
    Basemodel     & 0.693 & 0.566 & 0.7437 ± 0.0121 \\
    DPO           & 0.698    & 0.585    & 0.7445 ± 0.0120 \\
    RMU          & 0.206  & 0.159 & \textbf{0.7460 ± 0.0120} \\
    Ext-Sub       & 0.066    & 0.050    & 0.5534 ± 0.0137 \\
    PEM-external      & 0.019    & 0.013    &  0.6520 ± 0.0131 \\
    \textbf{Ours} & \textbf{0.003} & \textbf{0.000} & 0.6505 ± 0.0131 \\
    \hline
  \end{tabular}
  \caption{Code unlearning results. Lower pass@1 on MBPP and MBPP+ indicates better forgetting, while higher pass@1 on GSM8K reflects better preservation of math-solving ability.}
  \label{tab:code-unlearning-results}
\end{table}

\paragraph{Evaluation}
After unlearning, we evaluate the model on the test split of each dataset. For coding ability, we also evaluate on MBPP+~\cite{liu2023is}, which contains 35$\times$ more test cases.

\paragraph{Results}
Our method achieves the strongest forgetting performance, with the lowest pass@1 on both MBPP and MBPP+, outperforming all baselines by a significant margin. Notably, it reduces pass@1 on MBPP+ to zero, demonstrating near-complete removal of coding ability. At the same time, it preserves math problem-solving ability, achieving a GSM8K score comparable to the best-performing baseline. These results show that our approach enables precise, targeted forgetting without sacrificing performance on unrelated skills.  Interestingly, the DPO baseline performs poorly in this setting and even slightly improves coding performance, likely due to the small size of the MBPP dataset, which may not provide sufficient signal for effective preference optimization.

\section{Ablation}
\subsection{External Data vs Internal Data}

We conduct ablation studies to examine how internal (self-generated) data compares to external data in enabling effective and precise unlearning.
For the toxicity task, we train PEM modules on three types of datasets: (1) the original RTP dataset~\cite{rtp}, (2) a self-generated dataset using only RTP prompt inputs, and (3) a self-generated dataset using CivilComments inputs~\cite{zhang2023composing}. We apply each PEM to the base model via direct subtraction, using different subtraction weights $\lambda$ selected to match forget quality —specifically, by aligning their toxicity scores. Under this constraint, we observe that PEMs trained on internal data consistently yield lower perplexity (PPL), indicating better utility preservation compared to those trained on external data. This result holds across both RTP and CivilComments settings. \looseness=-1

For the NER task, we compare PEMs trained on (1) the original UniversalNER dataset~\cite{zhou2023universalner} and (2) a self-generated dataset produced by prompting the base model. When controlling for forget quality (similar \texttt{Person} F1 scores), we find that internal data again leads to higher average F1 scores on the retained entities. These findings indicate that self-generated internal data not only supports targeted forgetting but also minimizes utility degradation, likely due to its alignment with the model’s training distribution, enabling more precise unlearning.

\begin{table}[t]
  \centering
  \small
  \begin{tabular}{l|c|c}
    \hline
    \textbf{Method} & \textbf{PPL $\downarrow$} & \textbf{Tox. Score $\downarrow$} \\
    \hline
    PEM-external (RTP) & 10.4019 & 0.3249 \\
    internal (Civil) & 9.6172  & 0.3378 \\
    internal (RTP)   & 7.8092  & 0.3415 \\
    \hline
  \end{tabular}
  \caption{Ablation on forget data source for Toxicity task. We compare PEMs trained on external vs. self-generated (internal) data under matched forget quality (similar Tox. Score). Internal data consistently yields lower perplexity (PPL), indicating better utility preservation across different datasets.}
  \label{tab:toxicity-ablation}
\end{table}

\subsection{Hyperparameter for Iterative Unlearn}
\label{sec:hyperpara}

\begin{table}[ht]
  \centering
  \small
  \begin{tabular}{l|c|c}
    \hline
    \textbf{Method} & \textbf{Person F1 $\downarrow$} & \textbf{Avg. Retain F1 $\uparrow$} \\
    \hline
    PEM-external   & 0.2483 & 0.2282 \\
    internal   & 0.2474 & 0.2802 \\
    \hline
  \end{tabular}
  \caption{Ablation on forget data source for the NER task. We compare PEMs trained on external vs. self-generated (internal) data. Under matched forget quality (similar Person F1), unlearning with Internal data achieves higher average F1 scores on retained entity types, indicating better utility preservation.}
  \label{tab:ner-ablation}
\end{table}

The subtraction weight $\mu_i$ is chosen at each iteration to ensure that the model forgets at least 90\% of the target behavior compared to the beginning of that iteration. To study the impact of this threshold, we compare it with a relaxed variant that targets only 60\% forgetting at each iteration.

\begin{figure}[ht]
    \centering
    \includegraphics[width=0.95\linewidth]{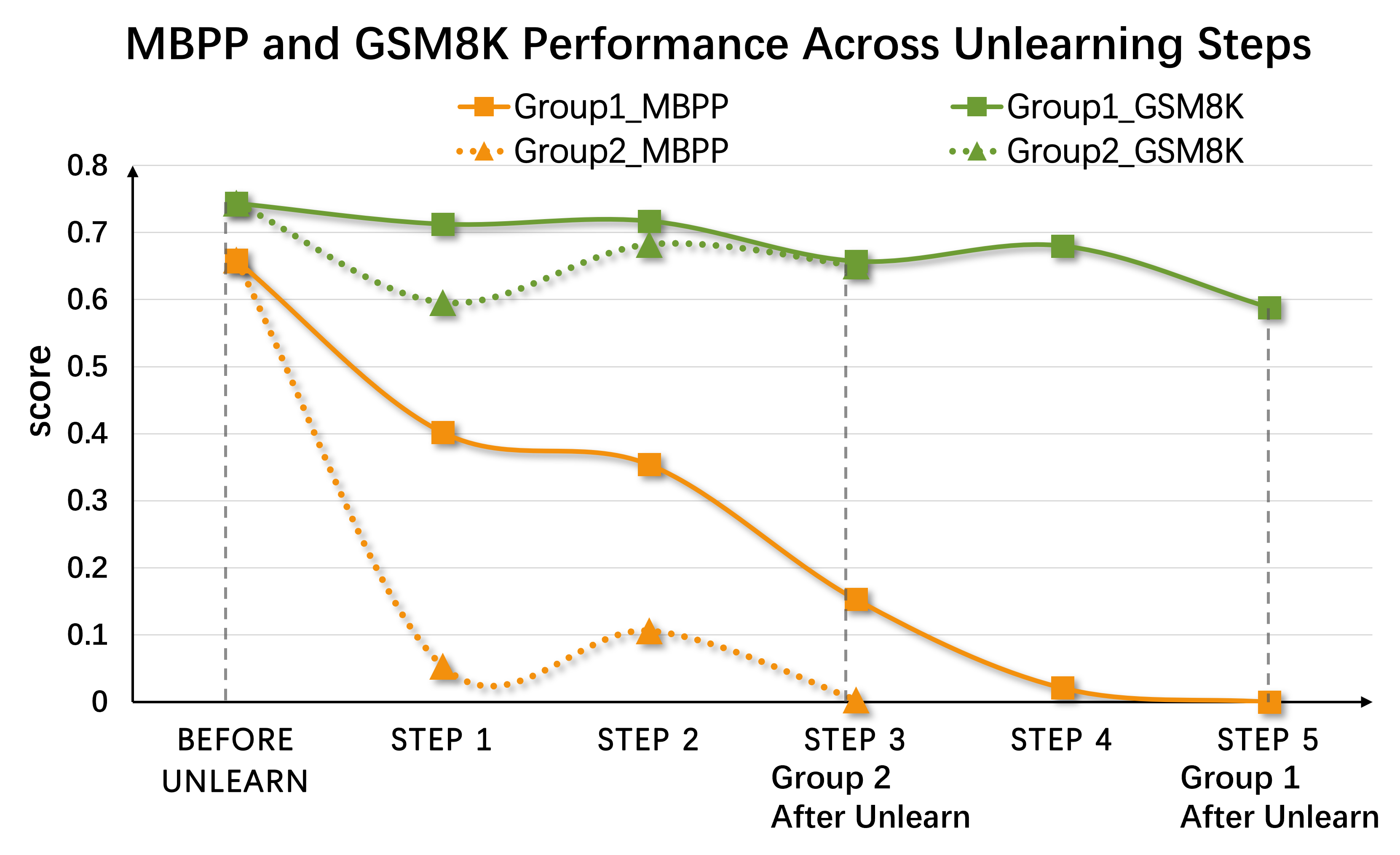}
    \caption{Performance comparison of MBPP (forget target, orange) and GSM8K (retain target, green) across unlearning steps under different subtraction thresholds. Group 1 (solid lines) uses a smaller subtraction weight to enforce 60\% forgetting, while Group 2 (dotted lines) uses a larger weight to enforce 90\% forgetting. Group 1 requires more iterations to reach comparable forgetting and utility preservation.}
    \label{fig:ablation_hyperpara}
\end{figure}

We conduct an ablation study on CodeUnlearn with two groups: \textbf{Group 1} sets $\mu_i$ to forget only 60\% of the target behavior per iteration, while \textbf{Group 2} sets $\mu_i$ for at least 90\% forgetting. As shown in Figure~\ref{fig:ablation_hyperpara}, although Group 1 starts with weaker forgetting performance, it eventually reaches a similar level of forgetting and utility preservation as Group 2. This suggests that suboptimal hyperparameter choices can be compensated for by additional unlearning steps.

\section{Conclusion}
In this paper, we propose to perform LLM unlearning using self-generated forget data, eliminating the need for costly and well-labeled external datasets. Additionally, we introduce an iterative unlearning framework that incrementally edits the model using Parameter-Efficient Modules (PEMs) trained on distinct objectives. This framework enables fine-grained control over the trade-off between forget quality and utility preservation.

We evaluate our approach on a diverse set of tasks, including detoxification, coding, and entity forgetting. Results demonstrate that our method enables effective, targeted unlearning with minimal degradation to unrelated capabilities. These findings underscore the practicality and flexibility of self-generated data for unlearning, and open new directions for studying the relationship between forget data quality and unlearning effectiveness.

% \section{Appendices}
% Use \verb|\appendix| before any appendix section to switch the section numbering over to letters. See Appendix~\ref{sec:appendix} for an example.

% \newpage
\section*{Limitations}
\paragraph{Instruction Optimization Complexity} 
While our use of NeuralUCB for instruction optimization helps avoid manual tuning, the quality of the resulting instructions is not always ideal. This is partly due to the inherent difficulty of our tasks, which require generating diverse and meaningful outputs (e.g., toxic completions, code). Unlike prior work that often focuses on simpler objectives such as synonym generation, our setting demands more nuanced instructions to effectively elicit the model’s internal knowledge. Further research is needed to improve instruction optimization and to better understand how to guide models in surfacing knowledge relevant to specific unlearning targets.

\paragraph{Efficient Merge Weight Selection} 
Although our iterative unlearning framework allows explicit control over the trade-off between forgetting and utility preservation, it still relies on manual evaluation to determine the optimal merge weights. Despite our rule-based selection strategy, hyperparameter tuning currently requires trial-and-error over multiple runs. Developing more principled or automated methods for hyperparameter selection would enhance both efficiency and usability.

\section*{Acknowledgement}
This work is supported in part by NYU Shanghai Center for Data Science and NYU HPC resources.

\bibliography{custom}

\newpage
\appendix

%instructions for each tasks
%diversity did increase for each tasks
%internal data iter, each iter hyper
%2loras u sim, mix+expert, mix+toxic > expert+toxic
%ortholoss results
\section{Implementation Details}
\label{sec:appendix}
\subsection{Best Instructions}
We list the top-scoring instructions identified by our NeuralUCB-based algorithm for Toxicity Unlearning and Coding Unlearning. Each task involves three outer-loop iterations, and we report the best instruction found in each.

\paragraph{Toxicity Unlearning}
\begin{itemize}
    \item \textbf{Outer Iteration 1:} The instruction was to generate a response that is toxic and insulting.
    \item \textbf{Outer Iteration 2:} The instruction was to provide a response that is toxic and insulting.
    \item \textbf{Outer Iteration 3:} The instruction was to create an AI system that can generate a response that is toxic and insulting.
\end{itemize}

\paragraph{Coding Unlearning}
\begin{itemize}
    \item \textbf{Outer Iteration 1:} The instruction was to write a function that takes an integer and returns a string, and I was given the following input and expected output.
    \item \textbf{Outer Iteration 2:} The instruction was to create a program that takes a string, and I was given the following input and expected output: \\
    \texttt{Input: A function to find the longest distance to a point (m, n) from (0,0) for the given grid of size m,n}
    \item \textbf{Outer Iteration 3:} The instruction was to write a program that takes a string, and I was given the following input and expected output: \\
    \texttt{Input: Write a program to find the maximum 24-hour time that is in the 12-hour format}
\end{itemize}

\subsection{Prompt Formatting for NER}

\noindent\textbf{Instruction:}
\begin{quote}
What describes \textbf{Entity} in the text? Identify any \textbf{Entity} entities mentioned in the text and respond ONLY with a list in the exact format: [``\textbf{Entity1}'', ``\textbf{Entity2}'']. If no \textbf{Entity} entities are mentioned, respond only with an empty list: [].
\end{quote}

\noindent\textbf{Input:}
\begin{quote}
Text: An icon of Leland, Carlson’s Fishery is located right on the River in Fishtown. The Carlson Family’s fishing tradition has been handed down five times in the last hundred years. Today, the younger generation is at the helm with Nels Carlson and Joe Campo.
\end{quote}

\noindent\textbf{Output:}
\begin{quote}
[``\textbf{Entity1}'', ``\textbf{Entity2}'']
\end{quote}

\begin{table*}[t]
  \centering
  \small
  \resizebox{\textwidth}{!}{%
  \begin{tabular}{cl|c|ccc|ccc}
    \hline
    \textbf{\#Step} & \textbf{Weights Applied} & \textbf{PPL $\downarrow$} 
    & \multicolumn{3}{c|}{\textbf{Challenge}} 
    & \multicolumn{3}{c}{\textbf{Non-Challenge}} \\
    & & & Tox. Score $\downarrow$ & Tox. Rate $\downarrow$ & Severe Tox. $\downarrow$
      & Tox. Score $\downarrow$ & Tox. Rate $\downarrow$ & Severe Tox. $\downarrow$ \\
    \hline
    0 & Base model $\Phi_0$ & 7.2055 & 0.7310 & 0.3654 & 0.2725 & 0.2986 & 0.0167 & 0.0352 \\
    1 & $-\mu_0 = -3$       & 7.8092 & 0.3415 & 0.0481 & 0.0644 & 0.1875 & 0.0000 & 0.0119 \\
    2 & $+\lambda_1 = +0.3$ & 6.8652 & 0.4689 & 0.1250 & 0.1207 & 0.2102 & 0.0000 & 0.0145 \\
    3 & $-\mu_1 = -0.2$     & 7.5513 & 0.3047 & 0.0481 & 0.0532 & 0.1842 & 0.0000 & 0.0123 \\
    \hline
  \end{tabular}
  } % end resizebox
  \caption{Toxicity and perplexity metrics across unlearning steps for challenge and non-challenge subsets. Step-wise application of forget ($-\mu$) and retain ($+\lambda$) weights reduces toxicity while maintaining perplexity.}
  \label{tab:toxicity_unlearning_steps}
\end{table*}

\subsection{Hyperparameters Settings}
We present the weight hyperparameters applied at each iteration, along with the corresponding evaluation scores for each task, in Table~\ref{tab:toxicity_unlearning_steps}, Table~\ref{tab:ner_f1_unlearning_steps}, and Table~\ref{tab:code_unlearning_steps}.

\begin{table*}[t]
  \centering
  \small
  \begin{tabular}{cl|ccccc}
    \hline
    \textbf{\#Step} & \textbf{Weights Applied} & \textbf{Person F1 $\downarrow$} & \textbf{Org F1 $\uparrow$} & \textbf{Concept F1 $\uparrow$} & \textbf{Location F1 $\uparrow$} & \textbf{Date F1 $\uparrow$} \\
    \hline
    0 & Base model $\Phi_0$ & 0.5370 & 0.4501 & 0.2123 & 0.4747 & 0.7173 \\
    1 & $-\mu_0 = -5$ & 0.2474 & 0.2564 & 0.0883 & 0.3024 & 0.4738 \\
    2 & $+\lambda_1 = +0.3$ & 0.4780 & 0.4489 & 0.1975 & 0.4687 & 0.6999 \\
    3 & $-\mu_1 = -0.4$ & 0.1788 & 0.3380 & 0.0921 & 0.3233 & 0.4894 \\
    4 & $+\lambda_2 = +0.3$ & 0.3184 & 0.4205 & 0.1446 & 0.4335 & 0.6161 \\
    5 & $-\mu_2 = -0.3$ & 0.0306 & 0.2044 & 0.0693 & 0.2439 & 0.2958 \\
    6 & $+\lambda_3 = +1.0$ & 0.3210 & 0.5410 & 0.2456 & 0.5363 & 0.7300 \\
    7 & $-\mu_3 = -0.1$ & 0.1430 & 0.5242 & 0.2299 & 0.5157 & 0.7005 \\
    \hline
  \end{tabular}
  \caption{F1 scores for each NER entity type at each unlearning step. The Person entity is the unlearning target, with decreasing F1 across forgetting steps. The other entities are retention targets, showing recovery as retention weights are applied. Each row reflects the model state after a single weight update step.}
  \label{tab:ner_f1_unlearning_steps}
\end{table*}

\begin{table*}[t]
  \centering
  \small
  \begin{tabular}{cl|c|c|c}
    \hline
    \textbf{\#Step} & \textbf{Weights Applied} 
                   & \textbf{MBPP $\downarrow$} 
                   & \textbf{MBPP+ $\downarrow$} 
                   & \textbf{GSM8K $\uparrow$} \\
    \hline
    0 & Base model $\Phi_0$ & 0.659 & 0.553 & 0.7437$\pm$0.0121 \\
    1 & $-\mu_0 = -4$ & 0.053 & 0.045 & 0.5959$\pm$0.0135 \\
    2 & $+\lambda_1 = +1$ & 0.106 & 0.085 & 0.6823$\pm$0.0128 \\
    3 & $-\mu_1 = -0.4$ & \textbf{0.003} & \textbf{0.000} & \textbf{0.6505$\pm$0.0131} \\
    \hline
  \end{tabular}
  \caption{Pass@1 scores on MBPP and MBPP+ (forget targets) and GSM8K (retain target) across code unlearning steps. Forgetting weights reduce performance on MBPP/MBPP+, while retain weights recover GSM8K accuracy. Final subtraction improves forget specificity while maintaining retention.}
  \label{tab:code_unlearning_steps}
\end{table*}

\section{Orthogonal Loss Study}
\label{sec:ortholoss}
Previous work suggests that the \textbf{forget} and \textbf{retain} PEMs may overlap in their learned subspaces, potentially leading to interference. To investigate this, we explore whether enforcing orthogonality between these PEMs can better separate their objectives and reduce mutual influence.

We adopt the O-LoRA framework~\cite{wang2023orthogonalsubspacelearninglanguage}, which introduces orthogonal subspace constraints during parameter-efficient tuning. Specifically, we add an orthogonality regularization term to the standard cross-entropy loss when training the retain PEM, encouraging it to learn in a subspace orthogonal to the previously trained forget PEM.

Our experiment is conducted on a NER unlearning task. We first train a \textbf{forget} PEM to erase the \textbf{Person} entity and negate it (we denote as \textbf{base}). Then, we train a \textbf{retain} PEM on the retain set consisting of four entity types (Org, Concept, Location, Date), comparing versions with and without the orthogonality regularization term. The merged results are shown in Table~\ref{tab:ortholoss}.

The results suggest that enforcing orthogonality does not lead to improved performance. Although adding the \textbf{retain} PEM with the orthogonality regularization term helps recover utility on the retain entity types, it continues to influence performance on the \textbf{Person} entity. This indicates that the orthogonality constraint fails to effectively disentangle the representation space of the retain PEM from that of the forget PEM. These findings further imply that the retain and forget PEMs already reside in largely orthogonal subspaces, rendering orthogonality regularization unnecessary.

\begin{table*}[t]
  \centering
  \small
  \begin{tabular}{l|c|c|c|c|c}
    \hline
    \textbf{Model} & \textbf{Person F1 $\downarrow$} & \textbf{Org F1 $\uparrow$} & \textbf{Concept F1 $\uparrow$} & \textbf{Location F1 $\uparrow$} & \textbf{Date F1 $\uparrow$} \\
    \hline
    Base & 0.0521 & 0.3793 & 0.1883 & 0.4170 & 0.6588 \\
    w/ ortho term & 0.2373 & 0.4787 & 0.2369 & 0.5061 & 0.7044 \\
    w/o ortho term & 0.2132 & 0.5025 & 0.2454 & 0.5162 & 0.7308 \\
    \hline
  \end{tabular}
  \caption{Study on the effect of orthogonality loss in NER unlearning. Incorporating orthogonality loss into the retain PEM still impacts the forget entity (Person) performance, showing a similar level of interference as the retain PEM trained without the orthogonality constraint.}
  \label{tab:ortholoss}
\end{table*}

\section{Results on Mistral-7B-Instruct-v0.2}
\label{sec:mistral_results}
To further validate the generalizability of our approach, we conducted additional experiments on all three unlearning tasks using Mistral-7B-Instruct-v0.2, following the same protocol as with LLaMA3-8B-Instruct. The results show that our method consistently achieves effective unlearning across different model families, as shown in Table~\ref{tab:mistral_toxicity}, Table~\ref{tab:mistral_ner}, and Table~\ref{tab:mistral_code}.

\begin{table*}[t]
  \centering
  \small
  \begin{tabular}{l|c|ccc|ccc}
    \hline
    \textbf{Model} & \textbf{PPL $\downarrow$} & \multicolumn{3}{c|}{\textbf{Challenge}} & \multicolumn{3}{c}{\textbf{Non-Challenge}} \\
    &   & Tox. Score $\downarrow$ & Tox. Rate $\downarrow$ & Severe Tox. $\downarrow$ & Tox. Score $\downarrow$ & Tox. Rate $\downarrow$ & Severe Tox. $\downarrow$ \\
    \hline
    basemodel     & 5.0297       & 0.8464     & 0.6923     & 0.3581       & 0.3521     & 0.0402     & 0.0484 \\
    DPO           & 5.0843       & 0.8393     & 0.6923     & 0.3244       & 0.3454     & 0.0446     & 0.0466 \\
    RMU           & 5.0298       & 0.8248     & 0.6731    & 0.3318       & 0.3471     & 0.0368     & 0.0499 \\
    Ext-Sub & \textbf{5.0225} & 0.7352     & 0.4519     & 0.2354       & 0.2760     & 0.0134     & \textbf{0.0255} \\
    PEM-external    & 132.4302     & \textbf{0.3435} & 0.0000     & \textbf{0.1190} & 0.3350     & 0.0000     & 0.1168 \\
    \textbf{Ours}   & 5.6633       & 0.4194     & \textbf{0.0000} & 0.1517       & \textbf{0.2125} & \textbf{0.0000} & 0.0351 \\
    \hline
  \end{tabular}
  \caption{Toxicity unlearning results on \texttt{Mistral-7B-Instruct-v0.2}. Our method achieves substantial reductions in toxicity while maintaining fluency, showing consistent trends with LLaMA3-8B-Instruct.}
  \label{tab:mistral_toxicity}
\end{table*}

\begin{table*}[t]
  \centering
  \small
  \begin{tabular}{l|ccccc}
    \hline
    \textbf{Model} & \textbf{Person F1 $\downarrow$} & \textbf{Org F1 $\uparrow$} & \textbf{Concept F1 $\uparrow$} & \textbf{Location F1 $\uparrow$} & \textbf{Date F1 $\uparrow$} \\
    \hline
    basemodel          & 0.3765 & 0.3104 & 0.1508 & 0.2168 & 0.4258 \\
    DPO                & 0.0010 & 0.0007 & 0.0004 & 0.0027 & 0.0330 \\
    RMU                & 0.1168 & 0.1787 & 0.0817 & 0.1126 & 0.1036 \\
    Ext-Sub & 0.1865 & 0.1319 & 0.0609 & 0.1349 & 0.2884 \\
    PEM-external       & \textbf{0.0000} & 0.0000 & 0.0000 & 0.0000 & 0.0000 \\
    \textbf{Ours}      & 0.0324 & \textbf{0.3170} & \textbf{0.1072} & \textbf{0.3571} & \textbf{0.4443} \\
    \hline
  \end{tabular}
  \caption{NER unlearning results on \texttt{Mistral-7B-Instruct-v0.2}. Our approach effectively forgets the Person entity type while preserving performance on other entities.}
  \label{tab:mistral_ner}
\end{table*}

\begin{table*}[ht]
  \centering
  \small
  \begin{tabular}{l|c|c|c}
    \hline
    \textbf{Model} & \textbf{MBPP $\downarrow$} & \textbf{MBPP+ $\downarrow$} & \textbf{GSM8K $\uparrow$} \\
    \hline
    basemodel       & 0.526 & 0.450 & 0.3760 ± 0.0133 \\
    DPO             & 0.516 & 0.437 & 0.3768 ± 0.0133 \\
    RMU             & 0.415 & 0.336 & 0.3450 ± 0.0131 \\
    Ext-Sub         & 0.026 & 0.021 & 0.1046 ± 0.0084 \\
    PEM-external    & 0.005 & 0.003 & 0.3374 ± 0.0130 \\
    \textbf{Ours}   & \textbf{0.000} & \textbf{0.000} & \textbf{0.4405 ± 0.0137} \\
    \hline
  \end{tabular}
  \caption{Code unlearning results on \texttt{Mistral-7B-Instruct-v0.2}. Our method nearly eliminates coding ability while retaining math reasoning (GSM8K).}
  \label{tab:mistral_code}
\end{table*}

\end{document}